%% file: neurips_main.tex
\documentclass{article}
\usepackage[preprint]{neurips_2022}
\input{defpack}

\title{\underline{P}ositive \underline{U}nlabeled \underline{Con}trastive Learning}

\author{%
  Anish Acharya\thanks{work done while interning at Meta.} \\
  UT Austin\\
  \And
  Sujay Sanghavi\\
  UT Austin \& Amazon\\
  \And
  Li Jing\\
  Meta\\
  \And
  Bhargav Bhushanam\\
  Meta\\
  \And
  Dhruv Choudhary\\
  Meta\\
  \And
  Michael Rabbat\\
  McGill University \& Meta\\
  \And
  Inderjit Dhillon\\
  UT Austin \& Google
}

\begin{document}

\maketitle

\begin{abstract}
\input{assets/abstract}
\end{abstract}

\section{Introduction}
\input{assets/introduction/intro}
\section{Problem Setup.}
\input{assets/problem_setting}
\section{Positive Unlabeled NCE}
\input{assets/algo/algo.puNCE}
\section{Related Work}
\input{assets/rel_work/related}

\section{Experiments}
\input{assets/exp/exp}
\subsection{Contrastive Positive Unlabeled Learning}
\input{assets/exp/con_pu}
\subsection{Contrastive Positive Negative Unlabeled learning}
\input{assets/exp/con_pnu}

\section{Conclusions}
\label{section:conclusion}
\input{assets/conclusion}

\bibliographystyle{apalike}
\bibliography{bibs/pu_learning, bibs/contrastive, bibs/exp_setup}

\newpage
\appendix
\section{Appendix}
\input{Notes/appendix}
%\newpage
%\input{checklist}
% Optionally include extra information (complete proofs, additional experiments and plots) in the appendix.
% This section will often be part of the supplemental material.

\end{document}

%% file: defpack.tex
\usepackage[utf8]{inputenc} % allow utf-8 input
\usepackage[T1]{fontenc}    % use 8-bit T1 fonts
\usepackage{hyperref}       % hyperlinks
\usepackage{url}            % simple URL typesetting
\usepackage{booktabs}       % professional-quality tables
\usepackage{amsfonts}       % blackboard math symbols
\usepackage{nicefrac}       % compact symbols for 1/2, etc.
\usepackage{microtype}      % microtypography
\usepackage{xcolor}         % colors
\usepackage{wrapfig}
\usepackage{graphicx}
\usepackage[algo2e]{algorithm2e}
\usepackage{comment}
\usepackage{subfig}
\usepackage{enumitem}
\setlist{leftmargin=3mm}
\usepackage{multirow}
\usepackage{tcolorbox}
\definecolor{light-gray}{gray}{0.9}
\usepackage{listings}
\input{math_commands}

\usepackage{color, colortbl}
\newcommand{\net}{f}

\usepackage[font=footnotesize]{caption}
\hypersetup{backref=true,       
    pagebackref=true,               
    hyperindex=true,                
    colorlinks=true,                
    breaklinks=true,                
    urlcolor= black,                
    linkcolor= blue,                
    bookmarks=true,                 
    bookmarksopen=false,
    filecolor=black,
    citecolor=blue,
    linkbordercolor=blue
}
\usepackage{pbox}
\usepackage{xcolor, soul}
\sethlcolor{light-gray}

\definecolor{LightCyan}{rgb}{0.88,1,1}
\newcolumntype{a}{>{\columncolor{LightCyan}}l}

\definecolor{codegreen}{rgb}{0,0.6,0}
\definecolor{codegray}{rgb}{0.5,0.5,0.5}
\definecolor{codepurple}{rgb}{0.58,0,0.82}
\definecolor{backcolour}{rgb}{0.95,0.95,0.92}

\lstdefinestyle{mystyle}{
    backgroundcolor=\color{backcolour},   
    commentstyle=\color{codegreen},
    keywordstyle=\color{magenta},
    numberstyle=\tiny\color{codegray},
    stringstyle=\color{codepurple},
    basicstyle=\ttfamily\footnotesize,
    breakatwhitespace=false,         
    breaklines=true,                 
    captionpos=b,                    
    keepspaces=true,                 
    numbers=left,                    
    numbersep=5pt,                  
    showspaces=false,                
    showstringspaces=false,
    showtabs=false,                  
    tabsize=2
}

%% file: math_commands.tex
\usepackage{amsmath,amsthm,amssymb,amsfonts}
\usepackage{algorithm}
\usepackage{algorithmic}
\usepackage{comment}
\usepackage{bm}
\usepackage{pifont}

\theoremstyle{plain}

\newtheorem{assumption}{Assumption}[]

\def\1{\mathbf{1}}

\def\rx{{\textnormal{x}}}
\def\ry{{\textnormal{y}}}

\def\vv{{\mathbf{v}}}

\def\vx{{\mathbf{x}}}

\def\vz{{\mathbf{z}}}

\DeclareMathAlphabet{\mathsfit}{\encodingdefault}{\sfdefault}{m}{sl}
\SetMathAlphabet{\mathsfit}{bold}{\encodingdefault}{\sfdefault}{bx}{n}

\def\gD{{\mathcal{D}}}

\def\gL{{\mathcal{L}}}

\def\gX{{\mathcal{X}}}

\def\sI{{\mathbb{I}}}

\def\sN{{\mathbb{N}}}

\def\sP{{\mathbb{P}}}
\def\sQ{{\mathbb{Q}}}
\def\sR{{\mathbb{R}}}

\def\sU{{\mathbb{U}}}

%% file: assets/abstract.tex
Self-supervised pretraining on unlabeled data followed by supervised finetuning on labeled data is a popular paradigm for learning from limited labeled examples. 
In this paper, we investigate and extend this paradigm to the classical positive unlabeled (PU) setting - the weakly supervised task
of learning a binary classifier only using a few labeled positive examples and a set of unlabeled samples. We propose a novel PU learning objective positive unlabeled Noise Contrastive Estimation (puNCE) that leverages the available explicit (from labeled samples) and implicit (from unlabeled samples) supervision to learn useful representations from positive unlabeled input data. The underlying idea is to assign each training sample an individual weight; labeled positives are given unit weight; unlabeled samples are duplicated, one copy is labeled positive and the other as negative with weights $\pi$ and $(1-\pi)$ where $\pi$ denotes the class prior. 
%Our method achieves state-of-the-art performances on PU learning benchmarks, including binary image classification and personalized recommendation systems. 
Extensive experiments across vision and natural language tasks reveal that puNCE consistently improves over existing unsupervised and supervised contrastive baselines under limited supervision. 

%% file: assets/introduction/intro.tex
\input{figs/compare_pu_algos}
% We note that,  PU Learning can also be viewed as binary {\bf biased semi-supervised learning}.
% In semi supervised 
% Since, in PU Learning all available labeled samples belong to only one class whereas, in semi-supervised learning some labeled samples corresponding to each class is available along with unlabeled samples. 
%In this context, 
Learning from limited amount of labeled data is a longstanding challenge in modern machine learning.
Owing to its recent widespread success in both computer vision and natural language processing tasks~\citep{chen2020big, Grill2020BootstrapYO, Radford2021LearningTV, Gao2021SimCSESC, dai2015semi, radford2018improving} 
{\em self supervised pretraining followed by supervised finetuning} has become the de-facto approach to learn from limited supervision. 
%owing to its widespread success in representation learning for both computer vision and natural language processing tasks.

These approaches typically leverage unlabeled data in a task agnostic manner during pretraining and use the available label information during finetuning~\citep{hinton2006fast, bengio2006greedy, mikolov2013distributed, kiros2015skip, devlin2018bert, Zbontar2021BarlowTS, assran2020supervision}. 
% e.g. in natural language processing, it is common to first train a language model from unlabeled text (ex. Wikipedia) followed by task specific downstream finetuning~\citep{mikolov2013distributed, kiros2015skip, devlin2018bert}.  
%with great success across several computer vision and natural language tasks. This motivates us to investigate the {\em self supervised pretraining followed by supervised finetuning} paradigm for PU Learning. 
% For example, in natural language processing, it is common to first train a language model from unlabeled text (ex. Wikipedia), followed by task specific downstream finetuning. 
In this context, contrastive learning has emerged as one of the most promising approach
% where the network learns 
to learn useful representations from unlabeled data %in a self-supervised manner 
by encouraging them to be invariant to distortions. 
% 
% SupCon~\citep{khosla2020supervised, zhong2021neighborhood}, a supervised variant of infoNCE that considers multiple positive pairs from other samples belonging to the same class as anchor in addition to the augmented view. By leveraging available supervision SupCon was able to obtain equally (or even more) discriminative representations as fully supervised cross entropy objective. 
\input{figs/pu_learning_problem}

This paper investigates and extends contrastive representation learning to the Positive Unlabeled (PU) Learning~\citep{denis1998pac} setting -  
%Dating back to the works of~\citep{denis1998pac}, Positive Unlabeled (PU) Learning 
%refers to 
the {\em weakly supervised} task of learning a binary classifier in absence of any explicitly labeled negative examples i.e. {\em using an incomplete set of positive and a set of unlabeled samples}. 
{\em PU learning also implies learning under distribution shift since it involves generalization to unseen negative class}~\citep{garg2021mixture}.  
This setting arises naturally in several real world applications
where obtaining negative samples is either too resource intensive or improbable. For example, consider {\em personalized recommendation systems}~\citep{naumov2019deep} where the training data is typically extracted from user feedback. Since explicit user feedback (e.g. user ratings) is hard to obtain, most practical recommendation systems rely on implicit user feedback (e.g. browsing history)~\citep{kelly2003implicit} which usually indicates user's positive preference (e.g. if a user browses a product frequently or watched a movie then the user-item pair is labeled positive)~\citep{chen2021pu}. This setup has also been motivated by applications like gene and protein identification~\citep{elkan2008learning}, anomaly detection~\citep{blanchard2010semi} and matrix completion~\citep{hsieh2015pu}. 

While self supervised contrastive losses e.g. InfoNCE \citep{gutmann2010noise} have gained remarkable success in unsupervised representation learning; recent works~\citep{he2020momentum, kolesnikov2019revisiting} have pointed out that purely self-supervised approaches often produce inferior visual representation compared to fully supervised approaches. To address this issue, researchers ~\citep{khosla2020supervised, assran2020supervision, graf2021dissecting} proposed supervised variants of contrastive loss (SCL) that, by leveraging available supervision, is able to obtain equally (or even more) discriminative representations as fully supervised cross entropy objective. 
Motivated by these observations, {\em our goal is to design a contrastive loss leveraging the available weak supervision in an efficient manner to learn rich representation $g_B(\vx)$ from Positive Unlabeled dataset, which is able to discriminate between positive and negative classes}.
% However, SCL relies on explicit label information and 
% in many real world settings, only a small fraction of samples are labeled implying in these settings SCL is only able to marginally improve over unsupervised representation. 
% These results intuitively suggest that incorporating available supervision (potentially correlated side information) might be helpful in learning rich representations from input data.
%and thus being able to leverage available weak supervision might help learning embedding equally rich as fully supervised training.  
% To address this issue, ~\citep{khosla2020supervised} developed supervised contrastive loss (SCL) that, by leveraging available supervision, is able to obtain equally (or even more) discriminative representations as fully supervised cross entropy objective. 

To this end, we propose {\bf puNCE} (positive unlabeled Noise Contrastive Estimation) - {\em a novel contrastive training objective that
extends the standard self supervised infoNCE~\citep{gutmann2010noise, chen2020simple} loss to the positive unlabeled setting;
by incorporating available biased supervision}. 
Unlike recent supervised variants of infoNCE~\citep{khosla2020supervised, assran2020supervision, zhong2021neighborhood} which can only leverage explicit (strong) supervision (e.g in form of labeled data), puNCE is also able to leverage implicit (weak) supervision from the unlabeled data. The main idea is to use the fact that unlabeled data is distributed as a mixture of positive and negative marginals where the mixture proportion is given by class prior $\pi$
~\citep{elkan2008learning, niu2016theoretical, du2014analysis, elkan2001foundations}. puNCE treats each unlabeled sample as a positive example and a negative example with appropriate probabilities and further uses the available explicit label information (i.e. set of positive labeled samples) to pull together multiple positive samples in the embedding space~\citep{khosla2020supervised}.
Our experiments across PU and binary semi-supervised settings suggest that 
% even with small amount of weak supervision 
% can dramatically improve the quality of the resulting embedding compared to unsupervised contrastive training using infoNCE. 
{\em in settings with limited supervision puNCE can be particularly effective and often produce stronger representations than both infoNCE and its supervised variants}~\citep{khosla2020supervised, assran2020supervision}. Our experiments on standard PU learning benchmarks reveal that 
even with small amount of weak supervision, 
puNCE can dramatically improve the quality of the resulting embedding compared to unsupervised contrastive training using infoNCE. For example, on PU CIFAR10 with ResNet-18 puNCE achieves over 3\% improvement compared to infoNCE while using 20\% labeled data, and over 2\% improvement when using only 5\% labeled data.   
% as well as performance in the downstream PU Learning task. 
% puNCE often results in improved representation compared to 
We further show that the resulting PU Learning approach i.e. {\em contrastive training the encoder using puNCE, followed by freezing the encoder and training a linear layer on top using standard cost-sensitive PU loss}~\citep{elkan2008learning, du2014analysis, kiryo2017positive} results in significant improvement in generalization performance compared to current state-of-the-art (Figure~\ref{fig:compare_pu}, Table~\ref{tab:cifar10_exp}). For example, on PU CIFAR-10 benchmark with only 1k positive labeled samples (i.e. rest 49k training samples unlabeled), our contrastive approach improves over existing PU learning approaches by 8.9\% with a ResNet-18.

% We refer to the resulting algorithm as {\bf ConPU} (contrastive PU). 
Our contributions can be summarized as follows:
\setlength{\intextsep}{0pt}
\begin{itemize} \setlength \itemsep{0.03 em}
\item We investigate the limitation of general self-supervised pretraining and finetuning approach on the weakly supervised PU learning task. We observe that self supervised contrastive losses are unable to leverage available supervision and produce inferior representations compared to fully supervised approaches. The gains from existing supervised contrastive losses are also limited when amount of explicitly labeled data is small (Table~\ref{tab:compare_cls}).   
\item We propose a novel contrastive training objective  - puNCE (positive unlabeled Noise Contrastive Estimation), that extends contrastive loss to the positive unlabeled setting by incorporating the available biased supervision via appropriately re-weighting the samples.   
\item We perform experiments on three settings with limited supervision - (a) Positive Unlabeled Classification, (b) Binary semi-supervised (PNU) classification and (c) Contrastive few shot finetuning of pretrained language model. Across all the setting, puNCE consistently outperforms supervised and unsupervised contrastive losses and is especially powerful when available supervision is scarce. On PU classification task, contrastive pretraining with puNCE results in large improvement over current state-of-the-art PU learning algorithms (Table~\ref{tab:cifar10_exp}).
% Our method achieved state-of-the-art performances on several PU learning benchmarks, including binary image classification and personalized recommendation systems.
\end{itemize}

%% file: figs/compare_pu_algos.tex
\begin{wrapfigure}{r}{0.4\textwidth}
\vspace*{-2 em}
  \begin{center}
    \includegraphics[width=0.4\textwidth]{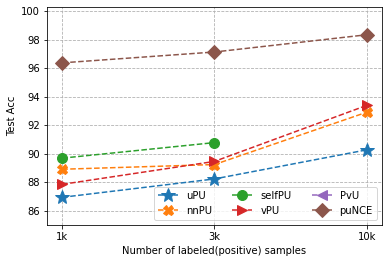}
  \end{center}
 \caption{{\footnotesize {\bf Comparison of PU Learning algorithms}: All-Conv trained on PU CIFAR (vehicle / animal) for varying number of labeled positive samples $n_P$ = 1k(2\%), 3k(6\%)}, 10k(20\%). The proposed contrastive loss {\bf puNCE} significantly improves over current PU baselines.}
 \label{fig:compare_pu}
% \vspace*{-1 em}
\end{wrapfigure}

%% file: figs/pu_learning_problem.tex
% \begin{wrapfigure}{r}{0.6\textwidth}
% \vspace*{-2 em}
%   \begin{center}
%     \includegraphics[width=0.6\textwidth]{img/PU_Problem.png}
%   \end{center}
%  \caption{{\footnotesize {\bf Comparison of PU Learning algorithms}: All-Conv trained on PU CIFAR (vehicle / animal) for varying number of labeled positive samples $n_P$ = 1k(2\%), 3k(6\%)}, 10k(20\%). The proposed contrastive loss {\bf puNCE} significantly improves over current PU baselines.}
% % \vspace*{-1 em}
% \end{wrapfigure}

% \begin{wrapfigure}{r}{0.7\textwidth}
\begin{figure}[]
%\vspace{-1 em}
\begin{center}
    \includegraphics[width=0.8\textwidth]{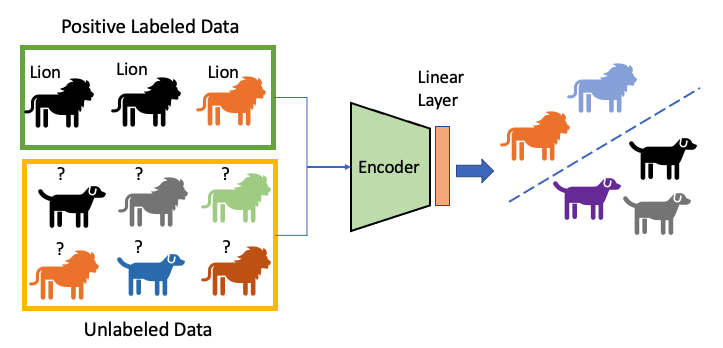}
\end{center}
\caption{\footnotesize {\bf Learning from Positive Unlabeled data.} 
No negative examples are labeled i.e. a classifier needs to be trained from an incomplete set of positive and a set of unlabeled samples such that it can discriminate between samples from positive and negative classes.}
\vspace{-1 em}
\label{fig:PU.problem}
\end{figure}
% \end{wrapfigure}

%% file: assets/problem_setting.tex
\subsection{Positive Unlabeled Learning.}
Let $\rx \in \sR^d$, $d \in \sN$ and $\ry \in \{\pm 1\}$ be the input and output random variables respectively and $p(\rx, \ry)$ be the true underlying joint density of $(\rx, \ry)$.
PU dataset $\gX_{PU}$ is composed of two independent sets of data, sampled uniformly at random from $p(\rx, \ry)$: an incomplete set $\gX_{P}$ of $n_p$ data-points with positive class and a set $\gX_{U}$ of $n_u$ unlabeled samples:  %i.e. $\gX_P$ and $\gX_U$ are sampled independently from the P and U marginals respectively~\citep{kiryo2017positive}.

\begin{equation}
\label{eq:pu_dataset}
\gX_P = \{\vx_i^P\}_{i=1}^{n_p} \sim p(\rx | \ry = 1);\; \gX_U = \{\vx_i^U\}_{i=1}^{n_u} \sim p(\rx);\; \gX_{PU} = \gX_P \cup \gX_U
\end{equation}

Since information about $y$ is unavailable for all samples, it is convenient to define an indicator random variable $s$ such that: {\bf s = 1 if $\vx$ is labeled and 0 otherwise}.
% \begin{flalign}
% s = 1 \text{if } \vx \text{ is labeled} 
% \begin{cases}
% 1 \; &\text{if } \vx \text{ is labeled}\\
% 0 \; &\text{o/w}
% \end{cases}
% \end{flalign}
Now viewing $\vx, y, s$ as random variables allows us to assume that there is some true underlying distribution $p(\vx, y, s)$ over triplets $(\vx, y, s)$ however, for each sample only $(\vx, s)$ is recorded. The definition of PU dataset~\eqref{eq:pu_dataset} immediately implies the following two results: 
\begin{flalign}
P(y = +1 | s = 1) = 1\;,\; p(s=1 | y=-1) = 0
\label{eq.pu.dataset.lemma.1}
\end{flalign}
% In contrast, in standard supervised setting (PN learning), negative samples $\gX_N = \{\vx_i^N\}_{i=1}^{n_n} \sim p(\rx| \ry=-1)$ are available instead of $\gX_U$. Let, $\ell:\sR\times\{\pm 1\}\to \sR$ denote the loss function and denote P marginal as $p_p(\rx) = p(\rx | \ry=1)$, N marginal $p_n(\rx) = p(\rx | \ry=-1)$ and U marginal $p(\rx)$, then the learner can be trained directly by minimizing the risk 
% $R(g) = \pi R_p^+(g) + (1- \pi)R_n^-(g)$ Where $R_p^+(g)$ and $R_n^-(g)$ are positive and negative risks respectively and are defined as:
% $
% R_p^+(g) = \E_{\rx \sim p_p}[\ell(\net(\rx), +1)]\;; \; R_n^-(g) = \E_{\rx \sim p_n}[\ell(\net(\rx), -1)].
% $ 
% However, in PU Learning since the negative samples are not provided, the negative risk cannot be estimated directly thus making it challenging.

\begin{assumption}[\bf Known Class Prior]
Throughout the paper, we assume that the class prior $\pi= p(y=+1)$ is either known or can be efficiently estimated from $\gX_{PU}$ via mixture proportion estimation algorithm~\citep{christoffel2016class, ramaswamy2016mixture,ivanov2020dedpul,garg2021mixture}.

Note that, this is a standard assumption made in PU Learning literature and is at the heart of most classical cost sensitive PU Learning algorithms~\citep{elkan2008learning, kiryo2017positive, du2014analysis, chen2020variational, niu2016theoretical}.  
\label{ass.pi}
\end{assumption}

\setlength{\fboxsep}{0.5pt}
\colorbox{light-gray}{\parbox{\columnwidth}{
The PU Learning task is thus to train a classifier $f:\sR^d \to \sR$ using only PU observations $(\vx, s)$ and knowledge of $\pi$, such that $f(\vx)$ is a close approximation of $p(y=+1|\vx)$. 
% In standard supervised setting (\underline{P}ositive \underline{N}egative(PN) learning), negative samples $\mathbf{\gX_N = \{\vx_i^N\}_{i=1}^{n_n} \sim p(\rx| \ry=-1)}$ are available instead of $\gX_U$. Thus, $f(\vx)$ can be trained directly via standard empirical risk minimization (ERM) framework. However, in PU Learning since the negative samples are not provided, the negative risk cannot be estimated directly.
Where, $\net$ is parameterized in terms of a linear layer $\vv \in \sR^k$ and feature extractor $g_B(\vx) \in \sR^{k \times d}$ i.e.
% \setlength{\fboxsep}{0pt}
% \colorbox{light-gray}{\parbox{\columnwidth}{
\begin{equation}
% \mathcolorbox{light-gray}{
    f_{\vv, B} = \vv^T g_B(\vx) : \sR^d \to \sR \;,\;  g_B(\vx) \in \sR^{k \times d} \;,\; \vv \in \sR^k
\label{eq:pu_learner}
\end{equation}}}

\subsection{Contrastive Representation Learning.}
% Recall that the goal of this paper is to design a contrastive loss that is able to leverage the available supervision in an efficient manner to learn rich representation $g_B(\vx)$ from Positive Unlabeled dataset that is able to discriminate between positive and negative classes.
The main idea of contrastive learning~\citep{sohn2016improved, wu2018improving} is to contrast semantically similar and dissimilar samples -  encouraging the representations of similar pairs to be close and that of the dissimilar pairs to be more orthogonal.   

In particular, at iteration $t$, given a randomly sampled batch $\gD_t = \{\vx_i \sim p(\vx)\}_{i=1}^b$ of b data points, the corresponding {\em multi-viewed batch} consists of 2b data points $\tilde{\gD_t} = \{\vx_i\}_{i=1}^{2b}$ where $\vx_{2i-1}$ and $\vx_{2i}$ are two views of $\vx_i \in \gD_t$ obtained via stochastic augmentations~\citep{chen2020self, tian2020contrastive}. Within the multi-viewed batch, consider an arbitrary augmented data point indexed by $i \in \sI \equiv \{1, \dots, 2b\}$ and let $a(i)$ be the index of the corresponding augmented sample originating from the same source sample. Denote the embedding of sample $\vx_i$ as $\vz_i = g_{B_t}(\vx_i)$, then the {\em self supervised contrastive loss} {\bf infoNCE}~\citep{gutmann2010noise, oord2018representation} is given as:
\begin{equation}
\small 
    \gL_{infoNCE} = \sum_{i \in \sI} \ell_{InfoNCE}^{(i)} = 
    \sum_{i \in \sI} \left[- \log \frac{\exp{(\vz_i} \boldsymbol \cdot \vz_{a(i)}/\tau)}{\sum_{k \in \sI \setminus \{i\}} \exp{(\vz_i\boldsymbol \cdot \vz_k / \tau)}}\right]
\label{eq:nt_xent}
\end{equation}
In this context $\vx_i$ is commonly referred to as the {\em \bf anchor}, $\vx_{a(i)}$ is called {\em \bf positive} and the other $2(b-1)$ samples are considered {\em \bf negatives}.
Simply put, InfoNCE loss pulls the anchor and the positive in the embedding space while pushing away anchor from negatives.

{\bf Projection Layer.} Rather than directly using the output of the encoder $g_B(\vx)$ to contrast samples, it is common practice~\citep{chen2020simple, Zbontar2021BarlowTS, Grill2020BootstrapYO, khosla2020supervised,  assran2020supervision} to feed the representation through a projection network $h_{\theta_{proj}}$ to obtain a lower dimensional representation before comparing the representations i.e. instead of using $\vz_i = g_{B_t}(\vx_i)$, in practice we consider $\vz_i = h_{\theta_{proj}}(g_{B_t}(\vx_i))$. {\em The projector is only used during optimization and discarded during downstream transfer task}. 

{\bf Contrastive learning with supervision.} 
Supervised Contrastive Learning (SCL)~\citep{khosla2020supervised, zhong2021neighborhood, graf2021dissecting, assran2020supervision} is a supervised variant of infoNCE that considers multiple positive pairs from other samples belonging to the same class as anchor in addition to the augmented view. in fully supervised setting, SCL can be computed as:

\begin{equation}
    \gL_{SCL} = \sum_{i \in \sI}\ell^{(i)}_{SCL}
    =-\sum_{i \in \sI} \frac{1}{|\sQ(i)|} \sum_{j\in \sQ(i)}\log \frac{\exp{(\vz_i} \boldsymbol \cdot \vz_j/\tau)}{\sum_{k \in \sI \setminus \{i\}} \exp{(\vz_i \boldsymbol \cdot \vz_k / \tau)}}
\label{eq:puNCE.P_risk}
\end{equation}
where, $\sQ(i) = \{ t \in \{\sI \setminus \{i\}\} : y_t = y_i\}$ denote the set of all samples in the current batch that belong to the same class as the anchor.

%% file: assets/algo/algo.puNCE.tex
%%%%%%%%%%%%%%%%%%%%%%%%%%%%%%%%%%%%%%%%%%%%%%%%%%%%%
\label{sec:puNCE}
%%%%%%%%%%%%%%%%%%%%%%%%%%%%%%%%%%%%%%%%%%%%%%%%%%%%%
% While self supervised contrastive losses e.g. InfoNCE \citep{gutmann2010noise} have gained remarkable success in unsupervised representation learning, \citep{he2020momentum, kolesnikov2019revisiting} have pointed out that purely self-supervised approaches often produce inferior visual representation compared to fully supervised approaches. These results intuitively suggest that incorporating available supervision (potentially correlated side information) might be crucial in learning rich representations from input data. 
% In this context, 
We propose {\bf puNCE} (positive unlabeled InfoNCE) that extends the self supervised InfoNCE loss to the PU learning setting. 
{\em The main idea of puNCE is to consider an unlabeled sample as positive with probability $\pi$ and negative with probability $(1 - \pi)$}. This follows from Bayes rule since we can write the marginal density of the unlabeled data as: 
% \setlength{\fboxsep}{0pt}
% \colorbox{light-gray}{\parbox{\columnwidth}{
\begin{flalign}
\label{eq.unlabeled_marginal}
    p(\vx) = \pi p(\vx | y=1) + (1- \pi) p(\vx | y = -1)
\end{flalign}
In particular, consider a {\bf labeled anchor} $\vx_i \in \gX_P$, the puNCE risk corresponding to $\vx_i$ is computed by pulling together normalized embeddings of all the available labeled samples in multi-viewed batch $\tilde{\gD}_t$ as opposed to InfoNCE which only considers $\vx_{a(i)}$ as positive. This holds, since all labeled samples are guaranteed to come from the same latent class as $\vx_i$~\eqref{eq.pu.dataset.lemma.1} . Given multi-view batch $\tilde{\gD_t}$, let $\sP \equiv \{k \in \sI : s_k = 1 \}$ denote the set of indices of all {\bf labeled samples} and $\sU \equiv \{k \in \sI : s_k = 0 \}$ indexes the {\bf unlabeled samples} i.e. $\sU \equiv \sI \setminus \sP$.
Then, puNCE empirical risk $\ell_P$ for the labeled samples is given as: 

\setlength{\fboxsep}{0pt}
\colorbox{light-gray}{\parbox{\columnwidth}{
\begin{equation}
    \ell_P = \sum_{i \in \sP}\ell^{(i)}_{puNCE}
    =-\sum_{i \in \sP} \frac{1}{|\sP|-1} \sum_{j\in \sP \setminus \{i\}}\log \frac{\exp{(\vz_i} \boldsymbol \cdot \vz_j/\tau)}{\sum_{k \in \sI \setminus \{i\}} \exp{(\vz_i \boldsymbol \cdot \vz_k / \tau)}}
\label{eq:puNCE.P_risk}
\end{equation}}}

On the other hand, {\bf unlabeled anchor} $\vx_i \in \gX_U$, is treated as positive example with probability $\pi=p(y=1|s=0)$ and as negative example with probability $(1-\pi)$. When considered positive ($y = +1$), all the labeled samples along with unlabeled augmentation $\vx_{a(i)}$ are used as positive pairs for $\vx_i$. Whereas, when it is considered negative ($y = -1$), since there are no available labeled negative samples puNCE treats only augmentation $\vx_{a(i)}$ as positive pair. The empirical risk on unlabeled samples $\ell_U$ is thus computed as:   

\setlength{\fboxsep}{0pt}
\colorbox{light-gray}{\parbox{\columnwidth}{
\begin{equation}
\small
\begin{aligned}
    \ell_U  = \sum_{i \in \sU}\ell^{(i)}_{puNCE}
    =- &\sum_{i \in \sU}\left[\frac{\pi}{|\sP|+1}\sum_{j\in \{\sP,a(i)\}} \log \frac{ \exp{(\vz_i} \boldsymbol \cdot \vz_j/\tau)}{\sum_{k \in \sI \setminus \{i\}} \exp{(\vz_i^t \boldsymbol \cdot \vz_k / \tau)}}\right. \\
    & \left.+ (1-\pi)\log \frac{\exp{(\vz_i} \boldsymbol \cdot \vz_{a(i)}/\tau)}{\sum_{k \in \sI \setminus \{i\}} \exp{(\vz_i \boldsymbol \cdot \vz_k / \tau)}}\right]
\end{aligned}
\label{eq:puNCE.U_risk}
\end{equation}}}

The first term of~\eqref{eq:puNCE.U_risk} denotes the loss incurred by the positive contribution of the unlabeled samples and the second term corresponds to the negative counterpart. Combining~\eqref{eq:puNCE.P_risk}, ~\eqref{eq:puNCE.U_risk} the puNCE empirical risk can be computed as:

\setlength{\fboxsep}{0pt}
\colorbox{light-gray}{\parbox{\columnwidth}{
\begin{equation}
\gL_{puNCE} = \frac{1}{|\tilde{D}_t|}\sum_{i\in \sI}\ell^{(i)}_{puNCE} = \frac{1}{2b}\left[\sum_{i\in \sP}\ell^{(i)}_{puNCE} + \sum_{i\in \sU}\ell^{(i)}_{puNCE}\right] = \frac{1}{2b}\left[\ell_P + \ell_U\right]
\label{eq.puNCE}
\end{equation}}}

This simply says that, in essence {\em puNCE assigns each training example an individual weight. In particular, all the labeled samples are given unit weight and the unlabeled samples are duplicated; one copy is labeled positive with weight $\pi$ and the other copy is labeled negative with weight $(1 - \pi)$}. 

\input{figs/pretrain_tsne}
\input{assets/algo/tab.compare_cls}

{\bf Extension to binary semi-supervised learning.}
PU Learning is closely related to semi-supervised learning. In semi-supervised learning labeled samples from both the classes along with unlabeled samples are available i.e. it contains three independent sets of samples: positive labeled, negative labeled and unlabeled samples (referred to as positive negative unlabeled (PNU) learning). A natural question to ask is if the idea of using implicit weak supervision from the unlabeled samples can also help in the general binary semi-supervised learning setting i.e. when the training data is limited but unbiased.
Let $\sP$, $\sN$ and $\sU$ denote the set of labeled positive, labeled negative and unlabeled samples respectively and $\sQ(i) = \{ t \in \{\sI \setminus \{i\}\} : y_t = y_i\}$ denote the set of all samples in the current batch that belong to the same class as the anchor and $\sP(i)=\{\sP,a(i)\}$, $\sN(i)=\{\sN,a(i)\}$, $\sI(i)=\sI \setminus i$. Then the puNCE loss in the PNU setting takes the following form: 

\setlength{\fboxsep}{0pt}
\colorbox{light-gray}{\parbox{\columnwidth}{
\begin{equation}
\begin{aligned}
    &\gL _{puNCE}^{PNU} = -\frac{1}{2b}\left[\sum_{i \in \sP\cup\sN} \frac{1}{|\sQ(i)|} \sum_{\sQ(i)}\log \frac{\exp{(\vz_i} \boldsymbol \cdot \vz_j/\tau)}{\sum_{k \in \sI \setminus \{i\}} \exp{(\vz_i \boldsymbol \cdot \vz_k / \tau)}}\right.\\
    +& \left.\sum_{i \in \sU} \left(\frac{\pi}{|\sP(i)|}\sum_{j\in \sP(i)} \log \frac{ \exp{(\vz_i} \boldsymbol \cdot \vz_j/\tau)}{\sum_{k \in \sI(i)} \exp{(\vz_i^t \boldsymbol \cdot \vz_k / \tau)}} + \frac{1 - \pi}{|\sN(i)|}\sum_{j\in \sN(i)} \log \frac{ \exp{(\vz_i} \boldsymbol \cdot \vz_j/\tau)}{\sum_{k \in \sI(i)} \exp{(\vz_i^t \boldsymbol \cdot \vz_k / \tau)}}\right)\right]
\end{aligned}
\end{equation}}}

{\bf Supervised and Unsupervised Setting}. It is straightforward to see that in the fully supervised setting puNCE reduces to SCL and in the unsupervised setting it is equivalent to infoNCE. 

%% file: figs/pretrain_tsne.tex
\begin{figure*}[t]
% \subfloat[\footnotesize{No PreTraining}]{\includegraphics[width=0.33\textwidth]{figs/TSNE/cifar10/data.png}}
\subfloat[\footnotesize{infoNCE}]{\includegraphics[width=0.33\textwidth]{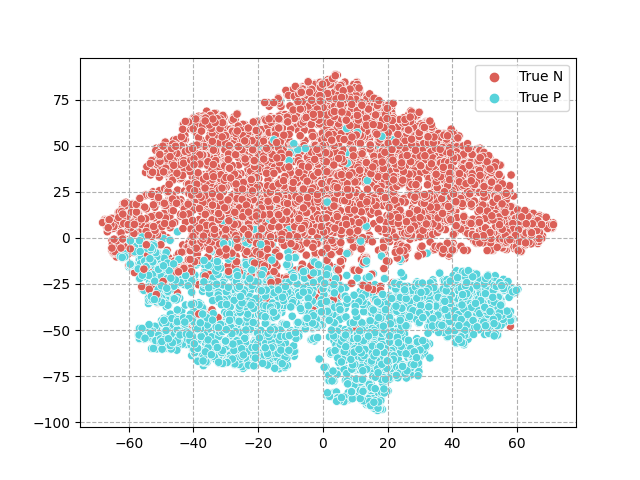}}
\subfloat[\footnotesize{puNCE ($n_p= 1k$)}]{\includegraphics[width=0.33\textwidth]{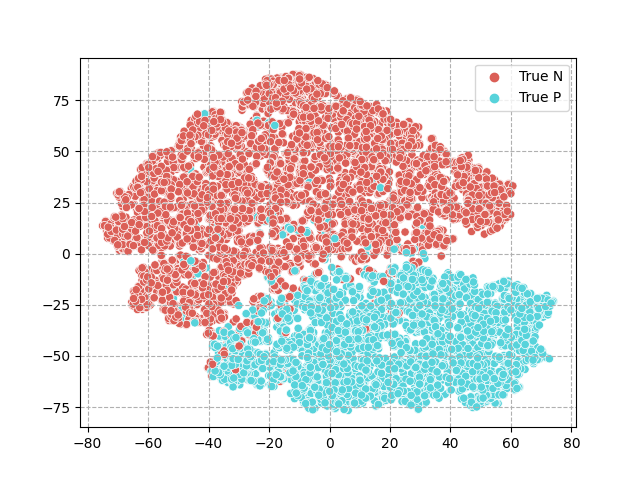}}
\subfloat[\footnotesize{puNCE ($n_P=3k$)}]{\includegraphics[width=0.33\textwidth]{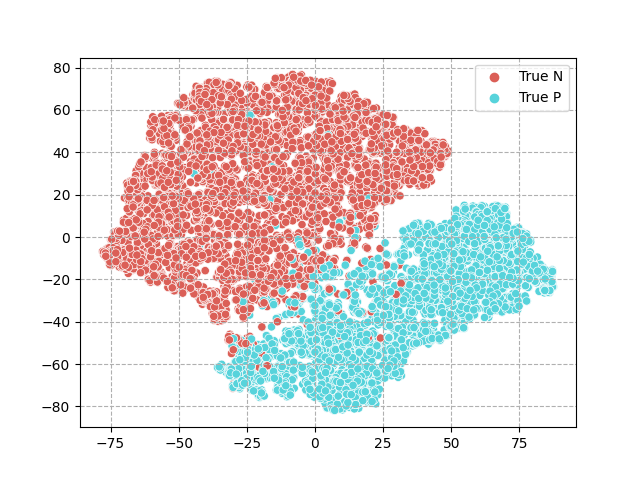}}
% \subfloat[\footnotesize{PU-InfoNCE ($10k$)}]{\includegraphics[width=0.25\textwidth]{img/TSNE/cifar10/tsne.pt.pu_con_debiased_pn.png}}
\caption{\footnotesize {\bf Performance with increased Label Information} t-SNE visualization of learned representations on CIFAR10. Classes are indicated by colors. Clearly the modified PUNCE objective leads to better class separation than InfoNCE. Further, we also note that the quality of the representation improves with increasing available positive labeled samples as expected intuitively.}
\label{figure:losses_tsne}
\end{figure*}

%% file: assets/algo/tab.compare_cls.tex
\begin{table*}[]
\centering
\footnotesize 
\resizebox{\textwidth}{!}{
\begin{tabular}{@{}ccccccc@{}}
\toprule
% \multirow{2}{*}{Dataset}                                
% & \multirow{2}{*}{$n_P$} 
% & infoNCE
% & supCon 
% & suNCEt
% & puNCE
Dataset  
& $g_B(\cdot)$
& $n_P$
& infoNCE
& DCL
& SCL
& puNCE
% \\
% &       
% & ~\citep{chen2020self}         
% & ~\citep{khosla2020supervised}     
% & ~\citep{assran2020supervision}       
% & {\bf (This work)}      
\\ \midrule \midrule 
\multirow{3}{1.75 cm}{\centering PU MNIST\\(odd/even)} 
& \multirow{3}{1.25 cm}{\centering MLP} 
& 1k    
& ${94.15\textcolor{gray}{\pm 0.15}}$         
& ${94.32\textcolor{gray}{\pm 0.42}}$        
& ${94.24\textcolor{gray}{\pm 0.09}}$        
& ${\bf 94.70\textcolor{gray}{\pm 0.19}}$     
\\ 
&
& 3k    
& ${94.84\textcolor{gray}{\pm 0.25}}$         
& ${95.09\textcolor{gray}{\pm 0.36}}$        
& ${95.82\textcolor{gray}{\pm 0.18}}$       
& ${\bf 96.01\textcolor{gray}{\pm 0.21}}$      
\\ 
&
& 10k    
& ${95.15\textcolor{gray}{\pm 0.05}}$         
& ${95.45\textcolor{gray}{\pm 0.08}}$       
& ${\bf 98.29\textcolor{gray}{\pm 0.08}}$        
& ${\bf 98.27\textcolor{gray}{\pm 0.11}}$      
\\\midrule 
\multirow{3}{2.5 cm}{\centering PU CIFAR \\ (animal / vehicle)} 
& \multirow{3}{2.5 cm}{\centering ResNet18} 
& 1k    
& ${96.33\textcolor{gray}{\pm 0.14}}$       
& ${96.21\textcolor{gray}{\pm 0.11}}$       
& ${97.42\textcolor{gray}{\pm 0.07}}$       
& ${\bf 97.59 \textcolor{gray}{\pm 0.17}}$     
\\ 
&
& 3k    
& ${96.51\textcolor{gray}{\pm 0.06}}$         
& ${96.49\textcolor{gray}{\pm 0.03}}$       
& ${97.66\textcolor{gray}{\pm 0.04}}$       
& ${\bf 97.87\textcolor{gray}{\pm 0.04}}$    
\\ 
&
& 10k    
& ${96.58\textcolor{gray}{\pm 0.04}}$       
& ${96.50\textcolor{gray}{\pm 0.02}}$        
& ${97.70\textcolor{gray}{\pm 0.07}}$       
& ${\bf 97.95\textcolor{gray}{\pm 0.02}}$    
\\\bottomrule                                          
\end{tabular}}
\caption{Linear Evaluation of different contrastive losses - infoNCE~\citep{chen2020simple}, DCL~\citep{chuang2020debiased}, SCL~\citep{khosla2020supervised} under Positive Unlabeled setting. puNCE is particularly effective when available supervision is limited. As we increase supervision, SCL starts to improve and become comparable. This interplay follows from the fact that they become equivalent in the fully supervised setting.}
\label{tab:compare_cls}
\end{table*}
% \begin{table*}[]
% \small 
% \centering
% \begin{tabular}{@{}ccccc@{}}
% \toprule
% Contrastive Loss
% & Label Information
% & $n_P=1k$ 
% & $n_P=3k$ 
% & $n_P=10k$ 
% \\ \midrule \midrule
% infoNCE~\citep{chen2020simple}
% & \color{red}\xmark
% & ${96.48\textcolor{gray}{\pm 0.25}}$          
% & $96.70\textcolor{gray}{\pm 0.05}$          
% & $96.72\textcolor{gray}{\pm 0.25}$  
% % \\ 
% % \textsc{DCL}~\citep{chuang2020debiased}
% % & \color{red}\xmark          
% % &      
% % &           
% % & 
% \\ 
% suNCEt~\citep{assran2020supervision}
% & \color{black}\cmark          
% &          
% &          
% & 
% \\
% SupCon~\citep{khosla2020supervised}
% & \color{black}\cmark          
% & ${\bf 96.68\textcolor{gray}{\pm 0.09}}$          
% & ${\bf 97.19\textcolor{gray}{\pm 0.12}}$          
% & 
% \\ 
% puNCE ({\bf This Work})
% & \color{black}\cmark          
% & %$96.37\textcolor{gray}{\pm 0.14}$          
% & %${97.13\textcolor{gray}{\pm 0.10}}$            
% & %${\bf 98.34\textcolor{gray}{\pm 0.09}}$ 
% \\ \bottomrule      
% \end{tabular}
% \caption{On CIFAR 10}
% \label{tab:ablation_contrastive_nP}
% \end{table*}

%% file: assets/rel_work/related.tex
{\bf PU Learning}.\;
Owing to its importance in several real world problems (e.g. recommendation), developing specialized learning algorithms for PU setting have received renewed impetus in the machine learning community. Most of the recent research in this area can be broadly categorized into two major class of algorithms - based on how they handle unlabeled samples. At a high level, the first class of algorithms~\citep{liu2002partially, liu2003building, bekker2020learning, luo2021pulns, chen2020self} tries to identify potential (with high confidence) negative, positive or both samples from the unlabeled pool of data; then perform traditional supervised learning together with the available labeled P data. The second set of algorithms~\citep{elkan2008learning, du2014analysis, kiryo2017positive} instead rely on finding appropriate weights for the positive and unlabeled samples and then performing standard cost sensitive training~\citep{elkan2001foundations}.
Our approach is closely related to the second set of methods as we treat each unlabeled sample as a convex combination of positive and negative samples and weight them accordingly. 

{\bf Contrastive Loss}.\; Self-supervised learning has demonstrated superior performances over supervised methods on various benchmarks. Joint-embedding methods~\citep{chen2020simple, Grill2020BootstrapYO, Zbontar2021BarlowTS, Caron2021EmergingPI} are one the most promising approach for self-supervised representation learning where the embeddings are trained to be invariant to distortions. To prevent trivial solutions, a popular method is to apply pulsive force between embeddings from different images, known as contrastive learning. Contrastive loss is shown to be useful in various domains, including natural language processing~\citep{Gao2021SimCSESC}, multimodal learning~\citep{Radford2021LearningTV}. Contrastive loss can also benefit supervised learning~\citep{khosla2020supervised}.

% However, in many real world scenarios the SCAR assumption might not be practical due to biases in observed positive labels~\citep{bekker2019beyond, su2021positive}. To deal with this some works also consider the following more relaxed assumptions.

% \begin{assumption}[\bf SAR: Selected At Random]

% \end{assumption}

% \begin{assumption}[\bf SNAR: Selected Not At Random]

% \end{assumption}

% \paragraph{Class Prior Estimation.}
% If the class prior $\pi$ of the unlabeled dataset is known, then the PU Learning problem~\eqref{equation:unlabeled_distribution} can be reduced to cost-sensitive learning~\citep{elkan2001foundations} over positive and unlabeled data. Let $\mC$ be the cost matrix then $\mC[i,j]$ i.e. the $(i,j)$ th entry denote the cost of predicting class $i$ when true class is $j$. 

%% file: assets/exp/exp.tex
In this section we state our experimental setup, 
present our empirical findings and discuss insights about how puNCE benefits learning from weak supervision. 
We compare puNCE with other popular contrastive losses including losses that can leverage available supervision. We also perform several ablations on puNCE and downstream PU classification. 
To ensure reproducibility, all the experiments are run with deterministic cuDNN back-end and repeated 5 times with different random seeds and the confidence intervals are noted.

{\bf Datasets.}
We evaluate puNCE on several common PU Learning image and text classification benchmarks consistent with recent literature~\citep{garg2021mixture, chen2020self, kiryo2017positive}.
In particular, for {\em vision benchmarks} we consider the following binary versions of MNIST and CIFAR10. (i) {\bf CIFAR10 (Vehicle / Animal)}: Animal images (bird, cat, deer, dog, frog, horse) are treated as negative and Vehicle images (airplane, automobile, ship, truck) are treated as positive. (ii) {\bf MNIST (Odd / Even)}: Separate the odd digits (1, 3, 5, 7, 9) from the even digits (0, 2, 4, 6, 8). We perform {\em text classification}, on {\bf SST2} sentiment dataset~\citep{socher2013recursive} where the goal is to classify positive and negative reviews. We simulate {\bf PU settings} from the binary data as follows: from the true positive samples in the training data we sample $n_P$ samples uniformly at random and label them positive while we mark the remaining training samples as unlabeled. 
For example, PU MNIST(Odd / Even) training dataset with $n_P = 1k$ consists of $1k$ labeled samples (positive) and $59k$ unlabeled ($29k$ true positive and $30k$ true negatives) samples. On the other hand to simulate {\bf binary semi-supervised PNU setting}, we instead randomly sample $n_l$ indices and keep the corresponding true label (labeled samples can contain both true positives and negatives), while all the other samples are marked as unlabeled ($s=0$). 
% Additionally we also perform experiments on two real world PU datasets - a) {\bf Alzheimer’s Disease brain image
% classification}. We perform experiments on ADNI dataset (b) Large scale deep learning recommendation system on Criteo Kaggle dataset. 
% \input{assets/exp/tab_dataset_details}

{\bf Learner Architecture.}
Recall that, our PU learner%~\eqref{eq:pu_learner} 
is parameterized in terms of an encoder $g_B(\cdot)$ with parameters $B$ and a linear layer with parameters $\vv$. On PU-CIFAR(Animal/Vehicle) we perform experiments with ResNet18~\citep{he2016deep} and All convolution net~\citep{springenberg2014striving} encoders. For PU-MNIST(Odd/Even) we train a MLP encoder with ReLU activation. For SST2 dataset we use a pretrained RoBERTa encoder~\citep{liu2019roberta}. 
% \paragraph{Effect of Number of Labeled Samples.} -- Since we already capture this in pre-training and also other PU papers have done this
% \paragraph{Robustness to Class Prior Estimation} 
%%%% MNIST 
%\input{assets/exp/tab_mnist_PU}

{\bf Contrastive Loss Baselines.}
We compare puNCE with several popular contrastive losses including unsupervised variants - InfoNCE~\citep{chen2020simple}, Debiased Contrastive Learning (DCL)~\citep{chuang2020debiased} as well as variants that can leverage explicit supervision - Supervised Contrastive Learning (abbreviated SupCon)~\citep{khosla2020supervised, assran2020supervision}. 
Following the setup of prior work on learning with limited supervision~\citep{chen2020big, assran2020supervision} we adopt SCL to the PU setting in the following manner: At iteration t, given an augmented batch $\tilde{\gD}_t$, on the labeled samples the contrastive loss is computed using SupCon loss while on the unlabeled samples in the augmented batch we compute the loss using standard infoNCE objective. 

\setlength{\fboxsep}{0pt}
\colorbox{light-gray}{\parbox{\columnwidth}{
\begin{equation}
    \gL_{SCL}^{PU}
    =-\frac{1}{2b}\left[\sum_{i \in \sP} \frac{1}{|\sP|-1}\sum_{j\in \sP \setminus \{i\}}\log \frac{\exp{(\vz_i} \boldsymbol \cdot \vz_j/\tau)}{\sum_{k \in \sI(i)} \exp{(\vz_i \boldsymbol \cdot \vz_k / \tau)}} + \sum_{i \in \sU}\ell_{infoNCE}^{(i)} \right]
\label{eq:supcon}
\end{equation}}}

{\bf PU Learning Baselines}. 
We compare the downstream PU classification performance of puNCE against
many related PU Learning algorithms at different levels of supervision ($n_P = 1k,3k,10k$). Our baselines include: PN~\citep{naumov2019deep, elkan2008learning} - considering all the unlabeled samples as negative and performing binary classification using standard loss (e.g. CE), cost sensitive approaches like PvU~\citep{elkan2001foundations}, uPU~\citep{du2014analysis}, nnPU~\citep{kiryo2017positive}, and other approaches like vPU~\citep{chen2020variational},SelfPU~\citep{chen2020self}. 

\input{assets/exp/tab_vision_PU}

%% file: assets/exp/tab_vision_PU.tex
%%% CIFAR 10 
\begin{table*}%[!htbp]
\footnotesize
\centering
\resizebox{\textwidth}{!}{
\begin{tabular}{@{}cccccc@{}}
\toprule
% \midrule
% \multicolumn{5}{c}{\bf ResNet-18 - PU CIFAR (vehicle/animal)}
% \\ \midrule
Dataset
& Encoder
& Algorithm  
%& $n_P$ = 500 
& $n_P$ = 1k 
& $n_P$ = 3k 
& $n_P$ = 10k
% & supervised
\\ \midrule\midrule
\multirow{10}{2 cm}{\centering PU CIFAR\\(vehicle/animal)} 
&\multirow{5}{1.75 cm}{\centering ResNet 18}
& PN~\citep{elkan2008learning} 
& $68.19 \textcolor{gray}{\pm 1.38}$ 
& $68.12 \textcolor{gray}{\pm 0.17}$
& $79.64 \textcolor{gray}{\pm 0.28}$
\\
&
& PvU~\citep{elkan2008learning} 
& $88.39 \textcolor{gray}{\pm 0.15}$  
& $90.16 \textcolor{gray}{\pm 0.19}$
& $93.54 \textcolor{gray}{\pm 0.21}$
\\
&
& uPU\citep{du2014analysis}
& $87.94\textcolor{gray}{\pm 0.54}$ 
& $89.53\textcolor{gray}{\pm 0.12}$   
& $92.37\textcolor{gray}{\pm 0.35}$
\\
&
& nnPU~\citep{kiryo2017positive}
&  $88.69\textcolor{gray}{\pm 0.08}$   
&  $89.46\textcolor{gray}{\pm 0.22}$ 
&  $91.58\textcolor{gray}{\pm 0.21}$
\\
&
& vPU\citep{chen2020variational} 
& $75.71\textcolor{gray}{\pm 0.31}$
& $83.64\textcolor{gray}{\pm 0.28}$
& $89.65\textcolor{gray}{\pm 0.19}$
% \\
% &
% & SelfPU~\citep{chen2020self} 
% & $89.68\textcolor{gray}{\pm 0.22}$  
% & $90.77\textcolor{gray}{\pm 0.21}$ 
% & --
% % \\ 
\\\cmidrule{3-6}
&
& puNCE(This work)
& ${\bf 97.59 \textcolor{gray}{\pm 0.17}}$ 
& ${\bf 97.87\textcolor{gray}{\pm 0.04}}$   
& ${\bf 97.95\textcolor{gray}{\pm 0.02}}$
\\
\cmidrule{2-6}
% \multirow{4}{1.75 cm}{\centering PU CIFAR\\(vehicle/animal)}
& \multirow{5}{1.75 cm}{\centering All Conv}
& PvU~\citep{elkan2008learning} 
%& $59.74 \textcolor{gray}{\pm 0.04}$ 
& $59.86\textcolor{gray}{\pm 0.15}$ 
& $61.20\textcolor{gray}{\pm 0.24}$ 
& $75.25\textcolor{gray}{\pm 0.79}$
\\
&
& uPU~\citep{du2014analysis}
%& $83.58\textcolor{gray}{\pm 0.18}$ 
& $86.93\textcolor{gray}{\pm 0.28}$ 
& $88.22\textcolor{gray}{\pm 0.83}$  
& $90.26\textcolor{gray}{\pm 1.60}$
\\
&
&nnPU~\citep{kiryo2017positive}
%& $87.34\textcolor{gray}{\pm 0.27}$  
& $88.90\textcolor{gray}{\pm 0.30}$   
& $89.23\textcolor{gray}{\pm 0.33}$  
& $92.90\textcolor{gray}{\pm 0.58}$
\\
&
&SelfPU~\citep{chen2020self} 
& $89.68\textcolor{gray}{\pm 0.22}$  
& $90.77\textcolor{gray}{\pm 0.21}$ 
& -
\\ 
&
& vPU~\citep{chen2020variational} 
& $87.84\textcolor{gray}{\pm 0.47}$  
& $89.45\textcolor{gray}{\pm 0.39}$
& $93.38\textcolor{gray}{\pm 0.02}$
\\
\cmidrule{3-6}
&
& puNCE (This work)
& ${\bf 96.37\textcolor{gray}{\pm 0.14}}$          
& ${\bf 97.13\textcolor{gray}{\pm 0.10}}$            
& ${\bf 98.34\textcolor{gray}{\pm 0.09}}$
\\
\midrule
\multirow{4}{1.75 cm}{\centering PU MNIST\\(Odd/even)}
% \multicolumn{5}{c}{MLP(4 Layer) - PU MNIST  (Odd vs Even)}
% \\\midrule 
& \multirow{5}{1.75 cm}{\centering MLP}
& PN~\citep{elkan2008learning}
&  $65.32 \textcolor{gray}{\pm 0.90}$ 
&  $94.16 \textcolor{gray}{\pm 0.26}$
&  $97.02 \textcolor{gray}{\pm 0.19}$
\\
&
& PvU\citep{elkan2008learning} 
& $91.10\textcolor{gray}{\pm 0.91}$
& $93.71\textcolor{gray}{\pm 0.91}$
& $95.31\textcolor{gray}{\pm 0.48}$
\\
&
& uPU\citep{du2014analysis}               
& $91.14\textcolor{gray}{\pm 0.87}$
& $93.84\textcolor{gray}{\pm 0.16}$     
& $96.43\textcolor{gray}{\pm 0.20}$
\\
&
& nnPU\citep{kiryo2017positive}              
& $91.83\textcolor{gray}{\pm 0.79}$ 
& $95.53\textcolor{gray}{\pm 0.29}$      
& $97.18\textcolor{gray}{\pm 0.12}$
\\ \cmidrule{3-6}
% & SelfPU~\citep{chen2020self}          
% & $94.21\textcolor{gray}{\pm 0.54}$  
% & $96\textcolor{gray}{\pm 0.29}$ 
% & 
% \\
%\rowcolor{light-gray}
&
& puNCE %(This Work) 
& ${\bf 94.70 \textcolor{gray}{\pm 0.19}}$ 
& ${\bf 96.01\textcolor{gray}{\pm 0.21}}$   
& ${\bf 98.27\textcolor{gray}{\pm 0.11}}$
\\
\bottomrule
\end{tabular}}
\caption{PU Computer Vision Benchmarks. We compare our contrastive PU Learning approach i.e. training encoder $g_B$ using puNCE, subsequently freezing the encoder and only training a linear layer on top. For all these experiments linear probing is done using nnPU loss.}
\label{tab:cifar10_exp}
\end{table*}
% \multicolumn{4}{c}{\bf LeNet(13 layer) - PU CIFAR (vehicle/animal)}
% \\ \midrule
% Algorithm  
% %& $n_P$ = 500 
% & $n_P$ = 1k 
% & $n_P$ = 3k 
% & $n_P$ = 10k 
% \\ \midrule
% PvU~\citep{elkan2008learning} 
% %& $59.74 \textcolor{gray}{\pm 0.04}$ 
% & $59.86\textcolor{gray}{\pm 0.15}$ 
% & $61.20\textcolor{gray}{\pm 0.24}$ 
% & $75.25\textcolor{gray}{\pm 0.79}$
% \\
% uPU~\citep{du2014analysis}
% %& $83.58\textcolor{gray}{\pm 0.18}$ 
% & $86.93\textcolor{gray}{\pm 0.28}$ 
% & $88.22\textcolor{gray}{\pm 0.83}$  
% & $90.26\textcolor{gray}{\pm 1.60}$
% \\
% nnPU~\citep{kiryo2017positive}
% %& $87.34\textcolor{gray}{\pm 0.27}$  
% & $88.90\textcolor{gray}{\pm 0.30}$   
% &  $89.23\textcolor{gray}{\pm 0.33}$  
% & $92.90\textcolor{gray}{\pm 0.58}$
% \\
% SelfPU~\citep{chen2020self} 

% %&   
% & $89.68\textcolor{gray}{\pm 0.22}$  
% & $90.77\textcolor{gray}{\pm 0.21}$ 
% & -
% \\ 
% \rowcolor{light-gray}
% ConPU ({\bf This work})
% & ${\bf 96.37\textcolor{gray}{\pm 0.14}}$          
% & ${\bf 97.13\textcolor{gray}{\pm 0.10}}$            
% & ${\bf 98.34\textcolor{gray}{\pm 0.09}}$
% \\
% \bottomrule \midrule
% \multicolumn{4}{c}{{\bf MLP(6 Layer) - PU MNIST (0-4 vs 5-9)}}
% \\\midrule 
% Algorithm  
% %& $n_P$ = 500 
% & $n_P$ = 1k 
% & $n_P$ = 3k 
% & $n_P$ = 10k 
% \\ \midrule
% PvU~\citep{elkan2008learning} 
% & 
% & 
% & 
% \\
% uPU~\citep{du2014analysis}               
% &  
% &      
% &
% \\
% nnPU~\citep{kiryo2017positive}              
% &  
% &      
% &
% \\
% SelfPU~\cite{chen2020self}            
% &   
% & 
% & 
% \\
% \rowcolor{light-gray}
% ConPU {\bf (This work)} 
% & 
% &      
% &
% \\
% \bottomrule \midrule

%% file: assets/exp/con_pu.tex
In this section we evaluate the performance of the proposed puNCE loss on downstream PU classification problem. 
As discussed before, our contrastive PU learning approach involves training encoder $g_B(\cdot)$ using contrastive loss, followed by training a linear layer on top using standard positive unlabeled loss (e.g. nnPU~\citep{kiryo2017positive}). 
For non contrastive classic PU baselines both the encoder ($B$) and the linear layer ($\vv$) are jointly trained using the PU loss.  
% \textcolor{red}{Note to self}
% Observations:
% \begin{itemize}
%     \item puNCE helps more in settings where available labeled data is low. Competitive with suNCEt
% \end{itemize}
% <contrastive pretraining> vary initial lr 1-.001 in steps on 10 and best is reported. 
% <finetuning> nnPU, uPU are less stable ~ use low lr 0.0001 - 0.0000001 , pretrained finetuning is stable can use large learning rate. 
\input{figs/punce_ablations}
Contrastive training is done using LARS optimizer~\citep{you2019large}, temperature set to 0.5. We used batch size 2048 for CIFAR10 experiments and 1024 for MNIST experiments. We used an initial learning rate of 0.01 with cosine annealing learning rate for 300 epochs on PU CIFAR10 and 200 epochs for PU MNIST.

{\bf Fine Tune vs Linear Probe.}

\input{assets/exp/tab.LPvFT}

After initializing the learner with the trained encoder parameters, we explore two popular transfer methods to transfer to the downstream Positive Unlabeled task - fine tuning (updating both $\vv$ and $B$), linear probing (updating $\vv$ while freezing the lower layers) using the PU training dataset. 

In standard ID setting i.e. when finetuning data and test data come from the same distribution, it is well known that finetuning results in better ID generalization performance than linear probing~\citep{he2020momentum, zhai2019large}. However, in the PU setting we observe that finetuning consistently under-performs linear probing. In fact, as the bias increases (i.e. for lower $n_P$), the gap becomes more significant (see Table 3). One possible way to explain this phenomenon is through feature distortion theorem~\citep{kumar2022fine}. Suppose, there exists an optimal $B_\ast$ and $v_\ast$ and suppose our representation quality is good such that our encoder parameters $B_0$ satisfies $\min_U\|B_0 - UB_\ast\|\leq \epsilon$ over all rotation matrices $U \in \sR^{k \times k}$ and $v$ is initialized randomly to $v_0$. Then we know that $v_0 v_0^T - B_0B_0^T = v_{ft} v_{ft}^T - B_{ft}B_{ft}^T$ where $v_{ft}, B_{ft}$ denote the parameters post finetuning~\citep{du2020few}. This suggest that since in PU setting the training data is biased, finetuning on it can result in significant feature distortion 
resulting in large error on data with large distribution shift. This is consistent with our observation, since as we reduce number of labeled samples i.e. we increase the bias and for small $n_P$, the distribution shift is large resulting in larger drop in performance. 
% that shows that fine tune can distort the pretrained features resulting in OOD error i.e. error on data having a different distribution than the data used for finetuning. Since in PU learning setting   
% in the downstream PU Learning task, the finetuning data is biased  
% After learning contrastive representation, we need to train the linear layer to adopt it to the downstream classification task. We experiment with several classic PU losses like PvU, uPU, nnPU as well as standard CE. 

{\bf Fine tune loss}. We experiment with different cost sensitive PU losses to during linear probing. We notice that they perform similarly with nnPU performing marginally better (Figure~\ref{fig:lp_nP}). 

{\bf Stability of PU Learning}. The classical cost-sensitive PU learning algorithms are notoriously unstable and tend to overfit the data. These methods need careful hyper parameter tuning and early stopping. Interestingly, we observe that when the model is initialized with parameters obtained from puNCE pretraining, the PU training becomes much more stable (Figure~\ref{fig:lp_stable}) suggesting that puNCE pretraining is also inherently robust.   

{\bf Effect of Number of Labeled Samples.}
To understand the benefit of increased supervision, we perform several experiments across different values of $n_P$. Our experiments (see Table~\ref{tab:compare_cls} and Figure~\ref{fig:cls_nP}) reveal that supervised approaches(SCL, puNCE) are consistently able to learn more powerful representation when supervision is available. The downstream performance of these approaches improve with increased supervision. {\em The results also suggest that puNCE can be particularly powerful when available supervision is small i.e. for smaller values of $n_P$ and results in significant gains over both unsupervised and supervised contrastive approaches further emphasizing the importance of leveraging implicit supervision from unlabeled samples}. 

%% file: figs/punce_ablations.tex
\begin{figure}[]
\subfloat[\footnotesize Contrastive pretraining]{\includegraphics[width=0.32\textwidth]{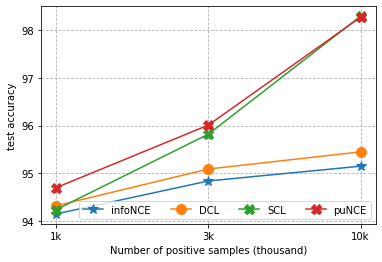} \label{fig:cls_nP}}
\subfloat[\footnotesize Linear Probing]{\includegraphics[width=0.33\textwidth]{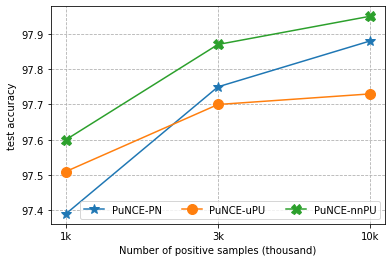}\label{fig:lp_nP}}
\subfloat[Training Stability]{\includegraphics[width=0.34\textwidth]{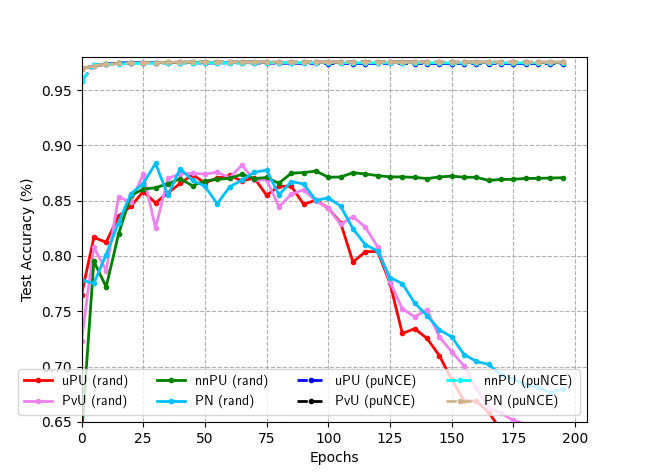}\label{fig:lp_stable}}
\caption{\footnotesize (ResNet18 - PU CIFAR10)
(\ref{fig:cls_nP}) Comparing different contrastive learning approaches on downstream PU classification task  across different amount of supervision ($n_P=1k,3k,10k$).
(\ref{fig:lp_nP}) Linear probing on top of puNCE features using standard supervised cross entropy (PN), nnPU~\citep{kiryo2017positive}, uPU~\citep{du2014analysis}.  
(\ref{fig:lp_stable})Test accuracy during training with puNCE initialization and random initialization of linear layer.  
}
\label{figure:punce_ablation}
\end{figure}

%% file: assets/exp/tab.LPvFT.tex
\begin{wraptable}{r}{0.5\textwidth}
\vspace{-0.5 em}
\caption{Comparing different transfer methods on training ResNet18 on PU CIFAR 10 - Linear probing performs consistently better than fine-tuning.}\label{tab:FTvLP}
\centering
    \footnotesize
    %\resizebox{\textwidth}{!}{
        \begin{tabular}{@{}ccc@{}}
        \toprule
        $n_P$
        & LP
        & FT
        \\\midrule\midrule
        1k
        & ${\bf 97.59\textcolor{gray}{\pm 0.17}}$ 
        & $92.96 \textcolor{gray}{\pm 0.19}$ (\textcolor{red}{$\downarrow 4.36$})
        \\
        3k
        & ${\bf97.87 \textcolor{gray}{\pm 0.04}}$
        & $95.83 \textcolor{gray}{\pm 0.15}$ (\textcolor{red}{$\downarrow 2.04$})
        \\
        10k
        & ${\bf97.95 \textcolor{gray}{\pm 0.02}}$
        & $97.41 \textcolor{gray}{\pm 0.09}$ (\textcolor{red}{$\downarrow 0.54$})
        \\
        \bottomrule
        \end{tabular}%}
    \vspace{0.5 em}
\end{wraptable}

%% file: assets/exp/con_pnu.tex
We further evaluate puNCE in the binary semi-supervised setting. Training data contains samples from both the classes and a set of unlabeled samples. In particular, we perform experiments when only 1\%, 5\% and 10\% of the data is available (Figure~\ref{tab:semi_sup}). 
It is important to note that, unlike PU Learning settings, here we perform downstream tuning only over the labeled data. We see similar trends as in PU Learning experiments - puNCE improves over infoNCE by 4.13\% and over SCL by 1.39\% when using only 1\% data on PU CIFAR10. With only 10\% data puNCE is able to achieve similar accuracy as a fully supervised baseline (with cross entropy) improving over semi-supervised adaptation of SCL~\citep{assran2020supervision} by 0.18\%.  
\input{assets/exp/tab_sem_sup}

\subsection{PNU few shot learning.}
Additionally, we also apply puNCE for pretrained language model finetuning (FT). In the few-shot learning setting, the goal is to FT a pretrained model on a downstream task using only a few labeled samples. FT with cross-entropy(CE) loss with only a few examples is highly unstable and can give very different results across different runs~\citep{dodge2020fine, zhang2018generalized}. To mitigate this, recent works~\citep{gunel2020supervised, chen2022dual} use contrastive loss in conjunction with CE i.e. $\gL = \lambda \gL_{CE} + (1 - \lambda) \gL_{CL}$. We closely follow the same setup, however, in addition to N labeled samples (N=10, 100, 1000), we also assume access to 500 unlabeled samples during finetuning. Note that for unlabeled samples, the CE term has no contribution. In this setting, we adopt a pretrained RoBERTa model for downstream SST2 binary sentiment classification and observe that puNCE results in 2.97\% (N=10), 0.93\%(N=100) and 0.47\% improvement over the previous state-of-the-art that uses SCL.   
\input{assets/exp/tab_SST}

%% file: assets/exp/tab_sem_sup.tex
\begin{table*}[]
\centering
\footnotesize 
% \setlength{\tabcolsep}{4 pt}
% \resizebox{\textwidth}{!}{
\begin{tabular}{@{}ccccccc@{}}
\toprule
Dataset  
& $g_B(\cdot)$
& Labeled
& infoNCE
% & DCL
& SCL
& puNCE
\\ \midrule \midrule 
\multirow{3}{2 cm}{\centering PNU CIFAR 10\\(vehicle/animal)} 
& \multirow{3}{1.25 cm}{\centering All Conv} 
& 1\%    
& $83.96\textcolor{gray}{\pm 2.45}$          
& $86.70\textcolor{gray}{\pm 0.93}$        
&  ${\bf 88.09\textcolor{gray}{\pm 1.88}}$    
\\ 
&
 &5\%    
& $93.92\textcolor{gray}{\pm 0.43}$         
& ${\bf 96.16\textcolor{gray}{\pm 0.06}}$       
& ${\bf 96.14\textcolor{gray}{\pm 0.34}}$    
\\
&
&10\%    
& $94.98\textcolor{gray}{\pm 0.34}$         
% & $93.22\textcolor{gray}{\pm 0.10}$       
& $96.39\textcolor{gray}{\pm 0.02}$        
& ${\bf 96.57\textcolor{gray}{\pm 0.11}}$    
\\\bottomrule
\end{tabular}
%}
\caption{Linear Evaluation of different contrastive losses under Semi-Supervised (PNU) setting - with 1\%, 5\% and 10\% labeled training data. puNCE proves to be superior than infoNCE~\citep{chen2020simple} and semi-supervised SCL~\citep{assran2020supervision} especially in low supervision regime.}
\label{tab:semi_sup}
\end{table*}

%% file: assets/exp/tab_SST.tex
\begin{table*}[]
\centering
\footnotesize 
% \setlength{\tabcolsep}{4 pt}
% \resizebox{\textwidth}{!}{
\begin{tabular}{@{}ccccccc@{}}
\toprule
Dataset  
& $g_B(\cdot)$
& Labeled
& CE
% & DCL
& CE + SCL
& CE + puNCE
\\ \midrule \midrule 
\multirow{3}{1.75 cm}{\centering PNU SST2} 
& \multirow{3}{1.25 cm}{\centering RoBERTa} 
& 20   
& $85.92\textcolor{gray}{\pm 2.11}$         
& $88.18\textcolor{gray}{\pm 3.30}$     
& ${\bf 91.15\textcolor{gray}{\pm 1.75}}$   
\\ 
&
 &100    
& $91.10\textcolor{gray}{\pm 1.36}$        
& ${92.83\textcolor{gray}{\pm 1.23}}$   
& ${\bf 93.76\textcolor{gray}{\pm 0.17}}$   
\\
&
&1000   
& $94.03\textcolor{gray}{\pm 0.64}$        
& $94.11\textcolor{gray}{\pm 0.53}$        
& ${\bf 94.58\textcolor{gray}{\pm 0.79}}$  
\\\bottomrule
\end{tabular}
%}
\caption{Few shot learning test results for N=10,100,1000. puNCE achieves significant gains over previous SCL based approach~\citep{gunel2020supervised} and standard CE baseline.}
\label{tab:few_shot}
\end{table*}

%% file: assets/conclusion.tex
In this work, we investigated the limitation of a general self-supervised pretraining and finetuning approach on weakly-supervised tasks.  We proposed a novel contrastive training objective puNCE (positive unlabeled Noise Contrastive Estimation), that extends contrastive loss to the positive unlabeled setting by incorporating available biased supervision. Our method achieved state-of-the-art performances on several PU learning and semi-supervised settings across vision and natural language processing and is particularly powerful when available supervision is limited.

%% file: Notes/appendix.tex
\subsection{Additional Experimental Details}

\subsubsection{Computing exact $\pi$}For our experiments we use the optimal $\pi$ i.e. probability of positive sample in unlabeled data exactly as follows: suppose $n_P$ positive samples are labeled and let $P^\ast$ and $N^\ast$ denote true number of positive and negatives. 
then we can readily use $\pi^\ast = \frac{P^\ast - n_P}{P^\ast + N^\ast - n_P}$.  

\subsubsection{Mixture Proportion Estimation}
In real settings when we do not have knowledge about the dataset exact $\pi$ cannot be computed and needs to be estimated - referred to as the Mixture proportion estimation (MPE) problem. Formally speaking, Mixture proportion estimation (MPE) refers to the task of estimating the weight of a component distribution in a mixture, given samples from the mixture (unlabeled data) and component (positive labeled data). We refer the reader to ~\citep{ramaswamy2016mixture, garg2021mixture, ivanov2020dedpul} for a detailed discussion on MPE algorithms. 

\subsubsection{MNIST Multi Layer Perceptron}
\begin{lstlisting}[language=python]
MLPContrastive(
  (backbone): Sequential(
    (0): Linear(in_features=784, out_features=5000, bias=True)
    (1): BatchNorm1d(5000, eps=1e-05, momentum=0.1, affine=True, track_running_stats=True)
    (2): ReLU(inplace=True)
    (3): Linear(in_features=5000, out_features=5000, bias=True)
    (4): BatchNorm1d(5000, eps=1e-05, momentum=0.1, affine=True, track_running_stats=True)
    (5): ReLU(inplace=True)
    (6): Linear(in_features=5000, out_features=50, bias=True)
    (7): BatchNorm1d(50, eps=1e-05, momentum=0.1, affine=True, track_running_stats=True)
    (8): ReLU(inplace=True)
  )
  (projector): Sequential(
    (0): Linear(in_features=50, out_features=300, bias=True)
    (1): BatchNorm1d(300, eps=1e-05, momentum=0.1, affine=True, track_running_stats=True)
    (2): ReLU(inplace=True)
    (3): Linear(in_features=300, out_features=50, bias=True)
  )
  (online_head): Linear(in_features=50, out_features=1, bias=True)
)
------------------
Num of Params = 29231151
\end{lstlisting}

\subsubsection{All Convolution N/w}
\begin{lstlisting}[language=python]
AllConv(
  (backbone_cnn): Sequential(
    (0): Conv2d(3, 96, kernel_size=(3, 3), stride=(1, 1), padding=(1, 1))
    (1): BatchNorm2d(96, eps=1e-05, momentum=0.1, affine=True, track_running_stats=True)
    (2): ReLU(inplace=True)
    (3): Conv2d(96, 96, kernel_size=(3, 3), stride=(1, 1), padding=(1, 1))
    (4): BatchNorm2d(96, eps=1e-05, momentum=0.1, affine=True, track_running_stats=True)
    (5): ReLU(inplace=True)
    (6): Conv2d(96, 96, kernel_size=(3, 3), stride=(2, 2), padding=(1, 1))
    (7): BatchNorm2d(96, eps=1e-05, momentum=0.1, affine=True, track_running_stats=True)
    (8): ReLU(inplace=True)
    (9): Conv2d(96, 192, kernel_size=(3, 3), stride=(1, 1), padding=(1, 1))
    (10): BatchNorm2d(192, eps=1e-05, momentum=0.1, affine=True, track_running_stats=True)
    (11): ReLU(inplace=True)
    (12): Conv2d(192, 192, kernel_size=(3, 3), stride=(1, 1), padding=(1, 1))
    (13): BatchNorm2d(192, eps=1e-05, momentum=0.1, affine=True, track_running_stats=True)
    (14): ReLU(inplace=True)
    (15): Conv2d(192, 192, kernel_size=(3, 3), stride=(2, 2), padding=(1, 1))
    (16): BatchNorm2d(192, eps=1e-05, momentum=0.1, affine=True, track_running_stats=True)
    (17): ReLU(inplace=True)
    (18): Conv2d(192, 192, kernel_size=(3, 3), stride=(1, 1), padding=(1, 1))
    (19): BatchNorm2d(192, eps=1e-05, momentum=0.1, affine=True, track_running_stats=True)
    (20): ReLU(inplace=True)
    (21): Conv2d(192, 192, kernel_size=(1, 1), stride=(1, 1))
    (22): BatchNorm2d(192, eps=1e-05, momentum=0.1, affine=True, track_running_stats=True)
    (23): ReLU(inplace=True)
    (24): Conv2d(192, 10, kernel_size=(1, 1), stride=(1, 1))
    (25): BatchNorm2d(10, eps=1e-05, momentum=0.1, affine=True, track_running_stats=True)
    (26): ReLU(inplace=True)
  )
  (backbone_lin): Sequential(
    (0): Linear(in_features=640, out_features=1000, bias=True)
    (1): ReLU(inplace=True)
    (2): Linear(in_features=1000, out_features=1000, bias=True)
    (3): ReLU(inplace=True)
  )
  (projector): Sequential(
    (0): Linear(in_features=1000, out_features=1000, bias=True)
    (1): BatchNorm1d(1000, eps=1e-05, momentum=0.1, affine=True, track_running_stats=True)
    (2): ReLU(inplace=True)
    (3): Linear(in_features=1000, out_features=1000, bias=True)
  )
  (online_head): Linear(in_features=1000, out_features=1, bias=True)
)
------------------
Num of Params = 5019255
\end{lstlisting}

\subsubsection{ResNet18}
\begin{lstlisting}[language=python]
ResNet18Contrastive(
  (backbone): Sequential(
    (0): Conv2d(3, 64, kernel_size=(3, 3), stride=(1, 1), padding=(1, 1), bias=False)
    (1): BatchNorm2d(64, eps=1e-05, momentum=0.1, affine=True, track_running_stats=True)
    (2): ReLU(inplace=True)
    (3): Sequential(
      (0): BasicBlock(
        (conv1): Conv2d(64, 64, kernel_size=(3, 3), stride=(1, 1), padding=(1, 1), bias=False)
        (bn1): BatchNorm2d(64, eps=1e-05, momentum=0.1, affine=True, track_running_stats=True)
        (relu): ReLU(inplace=True)
        (conv2): Conv2d(64, 64, kernel_size=(3, 3), stride=(1, 1), padding=(1, 1), bias=False)
        (bn2): BatchNorm2d(64, eps=1e-05, momentum=0.1, affine=True, track_running_stats=True)
      )
      (1): BasicBlock(
        (conv1): Conv2d(64, 64, kernel_size=(3, 3), stride=(1, 1), padding=(1, 1), bias=False)
        (bn1): BatchNorm2d(64, eps=1e-05, momentum=0.1, affine=True, track_running_stats=True)
        (relu): ReLU(inplace=True)
        (conv2): Conv2d(64, 64, kernel_size=(3, 3), stride=(1, 1), padding=(1, 1), bias=False)
        (bn2): BatchNorm2d(64, eps=1e-05, momentum=0.1, affine=True, track_running_stats=True)
      )
    )
    (4): Sequential(
      (0): BasicBlock(
        (conv1): Conv2d(64, 128, kernel_size=(3, 3), stride=(2, 2), padding=(1, 1), bias=False)
        (bn1): BatchNorm2d(128, eps=1e-05, momentum=0.1, affine=True, track_running_stats=True)
        (relu): ReLU(inplace=True)
        (conv2): Conv2d(128, 128, kernel_size=(3, 3), stride=(1, 1), padding=(1, 1), bias=False)
        (bn2): BatchNorm2d(128, eps=1e-05, momentum=0.1, affine=True, track_running_stats=True)
        (downsample): Sequential(
          (0): Conv2d(64, 128, kernel_size=(1, 1), stride=(2, 2), bias=False)
          (1): BatchNorm2d(128, eps=1e-05, momentum=0.1, affine=True, track_running_stats=True)
        )
      )
      (1): BasicBlock(
        (conv1): Conv2d(128, 128, kernel_size=(3, 3), stride=(1, 1), padding=(1, 1), bias=False)
        (bn1): BatchNorm2d(128, eps=1e-05, momentum=0.1, affine=True, track_running_stats=True)
        (relu): ReLU(inplace=True)
        (conv2): Conv2d(128, 128, kernel_size=(3, 3), stride=(1, 1), padding=(1, 1), bias=False)
        (bn2): BatchNorm2d(128, eps=1e-05, momentum=0.1, affine=True, track_running_stats=True)
      )
    )
    (5): Sequential(
      (0): BasicBlock(
        (conv1): Conv2d(128, 256, kernel_size=(3, 3), stride=(2, 2), padding=(1, 1), bias=False)
        (bn1): BatchNorm2d(256, eps=1e-05, momentum=0.1, affine=True, track_running_stats=True)
        (relu): ReLU(inplace=True)
        (conv2): Conv2d(256, 256, kernel_size=(3, 3), stride=(1, 1), padding=(1, 1), bias=False)
        (bn2): BatchNorm2d(256, eps=1e-05, momentum=0.1, affine=True, track_running_stats=True)
        (downsample): Sequential(
          (0): Conv2d(128, 256, kernel_size=(1, 1), stride=(2, 2), bias=False)
          (1): BatchNorm2d(256, eps=1e-05, momentum=0.1, affine=True, track_running_stats=True)
        )
      )
      (1): BasicBlock(
        (conv1): Conv2d(256, 256, kernel_size=(3, 3), stride=(1, 1), padding=(1, 1), bias=False)
        (bn1): BatchNorm2d(256, eps=1e-05, momentum=0.1, affine=True, track_running_stats=True)
        (relu): ReLU(inplace=True)
        (conv2): Conv2d(256, 256, kernel_size=(3, 3), stride=(1, 1), padding=(1, 1), bias=False)
        (bn2): BatchNorm2d(256, eps=1e-05, momentum=0.1, affine=True, track_running_stats=True)
      )
    )
    (6): Sequential(
      (0): BasicBlock(
        (conv1): Conv2d(256, 512, kernel_size=(3, 3), stride=(2, 2), padding=(1, 1), bias=False)
        (bn1): BatchNorm2d(512, eps=1e-05, momentum=0.1, affine=True, track_running_stats=True)
        (relu): ReLU(inplace=True)
        (conv2): Conv2d(512, 512, kernel_size=(3, 3), stride=(1, 1), padding=(1, 1), bias=False)
        (bn2): BatchNorm2d(512, eps=1e-05, momentum=0.1, affine=True, track_running_stats=True)
        (downsample): Sequential(
          (0): Conv2d(256, 512, kernel_size=(1, 1), stride=(2, 2), bias=False)
          (1): BatchNorm2d(512, eps=1e-05, momentum=0.1, affine=True, track_running_stats=True)
        )
      )
      (1): BasicBlock(
        (conv1): Conv2d(512, 512, kernel_size=(3, 3), stride=(1, 1), padding=(1, 1), bias=False)
        (bn1): BatchNorm2d(512, eps=1e-05, momentum=0.1, affine=True, track_running_stats=True)
        (relu): ReLU(inplace=True)
        (conv2): Conv2d(512, 512, kernel_size=(3, 3), stride=(1, 1), padding=(1, 1), bias=False)
        (bn2): BatchNorm2d(512, eps=1e-05, momentum=0.1, affine=True, track_running_stats=True)
      )
    )
    (7): AdaptiveAvgPool2d(output_size=(1, 1))
  )
  (projector): Sequential(
    (0): Linear(in_features=512, out_features=256, bias=False)
    (1): BatchNorm1d(256, eps=1e-05, momentum=0.1, affine=True, track_running_stats=True)
    (2): ReLU(inplace=True)
    (3): Linear(in_features=256, out_features=128, bias=True)
  )
  (online_head): Linear(in_features=512, out_features=1, bias=True)
)
------------------
Num of Params = 11333825
\end{lstlisting}